\documentclass[letterpaper]{article}
\usepackage[preprint]{style/aaai2027}
\usepackage{times}
\usepackage{helvet}
\usepackage{courier}
\usepackage[hyphens]{url}
\usepackage{style/natbib}
\usepackage{amsmath,amsfonts,bm}
\usepackage{graphicx}
\usepackage{booktabs}
\usepackage{array}
\usepackage{adjustbox}
\usepackage{xcolor}
\newcommand{\taut}{\tilde{\tau}}
\title{Wrong but Useful:\\Trajectory Value Beyond Answer Correctness in Multi-Agent Messages}
\newcommand{\resourcehref}[2]{%
  \leavevmode\pdfstartlink attr{/Border [0 0 0]} user{%
    /Subtype /Link /A << /S /URI /URI (#1) >>}%
  #2\pdfendlink}
\author{
Chih-Hsuan Yang\textsuperscript{1,*},
Anjir Ahmed Chowdhury\textsuperscript{2},
Cheng-Hau Yang\textsuperscript{1},\\
Weijian Zheng\textsuperscript{1},
Fernando Llorente\textsuperscript{3},
Xiaolong Ma\textsuperscript{1},
Xinyang Li\textsuperscript{2},\\
Eliu A. Huerta\textsuperscript{1,4},
Feng Yan\textsuperscript{2},
Ian T. Foster\textsuperscript{4,1},
Rajeev Thakur\textsuperscript{1}
}
\affiliations{
\textsuperscript{1}Argonne National Laboratory, Lemont, IL, USA\\
\textsuperscript{2}University of Houston, Houston, TX, USA\\
\textsuperscript{3}Brookhaven National Laboratory, Upton, NY, USA\\
\textsuperscript{4}University of Chicago, Chicago, IL, USA\\
\textsuperscript{*}Corresponding author: Chih-Hsuan Yang (\texttt{bellayang@anl.gov})\\[0.6em]
{\small
\resourcehref{https://chihhsuan-yang.github.io/AAAI_Wrong-but-Useful/}{Project website}
\quad\textbar\quad
\resourcehref{https://github.com/ChihHsuan-Yang/AAAI_Wrong-but-Useful}{Code}
\quad\textbar\quad
\resourcehref{https://huggingface.co/datasets/AgentsSci/AAAI_Wrong-but-Useful}{Dataset}
}
}

\begin{document}
\maketitle

\begin{abstract}
A solution can end with a wrong answer and still help another solver. In multi-agent reasoning, evaluating only a message's proposed answer can therefore miss its contribution to the receiver. We measure this contribution as \emph{trajectory value}: the change in a fixed integrator's probability of answering correctly when a cached message is present rather than removed. Controlled replay across five benchmarks and two model families separates message effects from identical-input variation. The strongest evidence is for Gemma: in a pre-registered fresh-call re-test, 34 of 73 selected wrong-answer messages pass individual confirmation, exceeding control-calibrated upper bounds on expected false confirmations of 11.4--20.8. The finding is benchmark-dependent and the conservative criterion no longer passes without SciBench. A blinded two-author audit supports the selected answer labels, while neutral replacements show that the helpful effect is not explained solely by message length and position. On held-out replay blocks of the same problems, a message-specific removal rule improves accuracy over the strongest tested correctness-only rule by 1.61 percentage points [1.01, 2.25]. Post hoc analysis localizes most of this gap to pools containing no correct proposal. Evidence for gpt-oss is weaker, and the tested predictors do not establish a practical way to identify useful errors in advance. The results support measuring answer correctness and receiver benefit separately, with value understood as conditional on the message pool, receiver and generation format.
\end{abstract}

\section{Introduction}
Suppose two solvers give the wrong answer to the same problem. One may offer a useful equation but make an arithmetic error; the other may offer a plausible argument that sends a reader in the wrong direction. Both fail an answer checker. Yet reading them need not have the same effect on the next solver. This distinction matters when a multi-agent system passes complete messages---answers together with reasoning---to an integrator that can combine, correct or reject their contents.

Our question is straightforward: \textbf{can a message with a wrong answer reliably help the receiving model, and what does its answer label fail to capture?} The word ``reliably'' matters. Removing a message and seeing the final answer change once is insufficient: identical model inputs can produce different outcomes even at temperature zero. A wrong-answer message can also appear helpful because it adds context length, because its answer was misgraded, or because all messages in its pool provide the same benefit. These alternatives require different checks.

We distinguish \emph{proposal correctness}, whether the message's own committed answer is correct, from \emph{trajectory value}, its effect on the integrator's success probability in a fixed pool. We cache messages, replay the same integrator with and without each one, and compare the resulting effects with identical-input noise and permutation references. We then test selected effects with fresh calls, inspect answer labels with human readers, and replace content with neutral text while keeping its slot and word count. We do not equate an incorrect final answer with incorrect reasoning throughout the message.

\paragraph{A concrete example.}
In a selected MaScQA rolling problem, a Gemma message gives a formula that evaluates to a reference-consistent 7.00\,mm, but commits to 1.73\,mm. All five proposed answers are wrong. In fresh calls the integrator is correct in 13 of 40 full-pool trials and 0 of 20 trials without this message. Removing a different wrong-answer message instead yields 19 correct answers in 20 trials. This post hoc example illustrates both possible directions of value; it does not establish a general mechanism or a population rate. Appendix J gives the messages, selection rule and uncertainty.

\paragraph{What the evidence supports.}
For Gemma, selected wrong-answer messages can help again in new calls, and their benefit survives several checks on labels and prompt footprint. Wrong messages can also harm, so retaining every wrong message is not the implication. A separate selection experiment asks whether knowing only which proposals are correct captures the useful distinctions. It does not fully do so: message-specific replay improves on the tested correctness-only rule. The largest gap appears when every available proposal is wrong. This is a diagnostic setting, not an assumption that systems should prefer wrong answers. If a correct answer is known, it can be used; our claim concerns useful information that can remain when such an answer is unavailable or when the receiver must evaluate the accompanying message.

The main contributions are: (1) a noise-calibrated measurement of correctness and trajectory value as distinct properties; (2) fresh-call and human-audit evidence for wrong-but-useful messages in a bounded Gemma setting; and (3) an analysis of where correctness-only selection loses information, together with empirical limits on mechanism and prediction. The evidence does not establish universal benefit across models or a deployable selector.

\section{Related Work}\label{sec:related}
\paragraph{Weak or wrong contributions can help on average.}
Mixture-of-Agents shows that an aggregator improves when shown other models' outputs, including weaker ones
\citep{wang2024moa}; mixing models can also hurt relative to the best single model \citep{li2025selfmoa}, and whether
multi-agent discussion improves on a strong single agent is itself contested \citep{wang2024rethinking,zhang2025madquestion}.
Aggregating full traces can recover correct solutions even when agents agree on a wrong answer
\citep{fadnavis2026beyond}, and majority voting explains much of the gain of debate
\citep{choi2025debate,du2023multiagent,chen2023reconcile}. Invalid chains of thought, wrong in-context labels and
failed trajectories still help models
\citep{min2022rethinking,wang2023towards,li2025structure,alazraki2025no,wang2024learningfromfailure,an2023lema}.
These findings motivate measuring whether a particular wrong-answer message helps a fixed receiver, and whether that effect survives replay noise.

\paragraph{Verifiers and process rewards.}
Process reward models score steps by their association with correct solutions
\citep{uesato2022process,lightman2023verify,wang2024mathshepherd,zhang2025lessonsprm}. Process advantage verifiers define
a step's value as its progress, the change in future success probability under a separate prover
\citep{setlur2024rewarding}, the estimand closest to ours; their unit is a step inside one solver's trajectory, ours a
complete cross-agent message. We do not test a process reward model.

\paragraph{Attribution and valuation.}
Influence functions, Data Shapley, datamodels and context attribution value inputs counterfactually
\citep{koh2017influence,ghorbani2019datashapley,ilyas2022datamodels,cohenwang2024contextcite,xie2024demoshapley}; in
retrieval, passage utility differs from relevance and depends on the generator
\citep{salemi2024erag,cuconasu2024power,zhang2025llmutility}. In multi-agent systems, C3 assigns \emph{message-level}
credit by counterfactual replay, characterises the variance of that credit, and reports that a critic trained on the
same rollouts recovers it poorly \citep{chen2026exact}. Agents that Matter compares leave-one-out with combinatorial
attribution at the \emph{agent} level and finds that introspective LLM judges do not approximate ablation
\citep{lu2026agents}. Causal Agent Replay attributes failures to \emph{steps} by re-execution \citep{shah2026causal};
see also failure attribution and learned approximations of contribution
\citep{zhang2025whowhen,zhang2025agentracer,cui2025introspecloo,dong2026bandmas,li2026e2explainer}.
\citet{yang2025whogets} propose a conceptual framework that turns system evaluation into signed agent- and message-level training signals. Our replay study measures receiver effects; it does not test the training gains of that framework.
Our contribution is the empirical distinction between a message's own answer correctness and its effect on a receiver, tested with correctness-by-value labels, noise-calibrated replay, fresh calls and label validation.

\paragraph{Message quality, uptake and collaboration value.}
\citet{yang2026precise} show that precise error detection by a reviewer need not lead to critique uptake or successful repair. At the protocol level, \citet{yang2026costaware} distinguish predicting a solver's failure from predicting which collaboration protocol will help. \citet{jiang2026confidence} separately audit confidence-based routing, calibration and whether an agent states the answer that won its turn. These studies motivate evaluating the receiving system's behavior separately from a message's apparent quality. Here the unit is one cached message: we test whether its presence changes a fixed integrator's success probability, including when its own answer is wrong.

\paragraph{Selection, pruning, distraction and answer order.}
Self-consistency, tree search and learned rankers keep several paths and select among them
\citep{wang2022selfconsistency,yao2023tot,jiang2023llmblender}; communication pruning and agent elimination remove
messages or agents to save cost \citep{zhang2025agentprune,wang2025agentdropout,liu2023dylan}. These methods optimise
a final answer; we measure the per-message target they approximate. Irrelevant context can distract LLM reasoning
\citep{shi2023distracted,yang2025distracting}, so we also compare a message with same-slot neutral text of equal word count. In coding agents, \citet{chen2026memory} distinguish stored state, delivered context, management work and task outcomes, showing why equal token budgets alone are insufficient for comparing memory policies. Our neutral-text control addresses the narrower question of whether a message's length and slot explain its measured benefit. Stated
reasoning need not reflect computation \citep{lanham2023measuring,turpin2023unfaithful}, and agents conform to peers'
wrong answers \citep{weng2025doaswedo,wynn2025talk}. That reasoning before answering changes the answer is the premise
of chain-of-thought prompting \citep{wei2022cot}; we use it only as a scope condition on which messages are labelled wrong (the format experiment).

\section{Measuring Trajectory Value}
\paragraph{Protocol and information boundaries.}
A recruiter sees a problem and assigns five problem-specific roles. Each hypothesizer sees the problem and one role, but no other message or reference answer. It produces a structured, answer-first message. The integrator sees the problem and a selected subset of these cached messages; it is free to synthesize rather than merely vote. Only the evaluator sees the reference answer. We use gpt-oss-120b (OSS) and gemma-4-31B-it (Gemma), with senders and integrators from the same family, on Omni-MATH-2, JEEBench, SciBench, LAB-Bench and MaScQA \citep{ballon2026omnimath2,arora2023jeebench,wang2023scibench,laurent2024labbench,zaki2023mascqa}. Both families use gpt-oss-120b for grading. Appendices A and L give decoding settings and prompt interfaces.

\paragraph{Two properties of a message.}
For problem $x$, let $P=\{m_1,\ldots,m_K\}$ be a fixed pool and $Y(P)\in\{0,1\}$ the graded correctness of the integrator's answer. Proposal correctness $c_i\in\{0,1\}$ scores the first committed answer in message $m_i$. Its trajectory value under removal is
\begin{equation}
\tau_i = \mathbb E[Y(P)]-\mathbb E[Y(P\setminus\{m_i\})].
\label{eq:value}
\end{equation}
The expectation is over repeated calls under the evaluation procedure. Positive value means that including the message improves success relative to its removal, in this pool and for this receiver; negative value means it reduces success. It is not an intrinsic score of the text or a guarantee that the receiver uses any particular reasoning step.

In a single paired observation, equal correctness outcomes give a zero delta. Across repeats, a confidence interval containing zero means that help or harm was not resolved, not that the true effect is exactly zero. A change in reasoning without a change in final correctness is invisible to that individual binary observation; it does not establish neutrality in expectation. We call repeatably positive and negative wrong-answer labels W$+$ and W$-$, with C$+$ and C$-$ for correct-answer messages.

\paragraph{Controlled replay and units.}
The controlled study uses 1,000 problems per family, 200 per benchmark. Each five-message pool has five replay blocks. A block contains two identical-input full-pool calls and five single-removal calls. The two full outcomes are averaged before subtracting each removal outcome. Replaying changes neither the cached messages nor the evaluator. Problems, messages, replay calls and inferred labels are different units: repeated calls to one message are not additional independent messages. Unless stated otherwise, intervals are 95\% problem-cluster bootstrap intervals; selection comparisons use paired resampling. Equal problem counts give benchmarks equal weight in the full controlled cohort, but selected subsets need not be balanced.

\paragraph{Noise and pool effects.}
Identical-input outcomes disagree in 2.1\% of Gemma blocks and 7.3\% of OSS blocks. We classify a message as stable helpful or harmful only when the bootstrap interval of its block-mean delta excludes zero. We compare label counts with an exact within-block permutation null that applies the same labeller. Permuting all seven outcomes tests the absence of any full-versus-removal effect; it includes effects common to a pool. A second null keeps full-pool outcomes fixed and permutes only the five removals, asking whether messages differ from their pool-mates. We report both and avoid interpreting their difference as new independent data. Original dispatch order was fixed; fresh-call experiments randomize arm order.

\paragraph{Content control.}
Let $F$, $R$ and $N$ denote mean correctness with the full pool, with the target removed and with its text replaced by neutral text in the same slot with the same word count. Then
\begin{equation}
F-R=(F-N)+(N-R).
\end{equation}
$F-N$ tests whether the message beats this neutral replacement; $N-R$ measures its remaining footprint. Word matching does not guarantee identical token counts or unchanged positions of every subsequent token. The Gemma control covers 969 of 1,000 problems with one fresh replay per arm; its missing problems limit generalization.

\paragraph{Study status.}
``Pre-registered'' denotes an internally frozen prospective protocol or analysis plan, rather than an external registry. The fresh re-test and format protocols were hashed before their data were collected; timing and records differ for other studies. ``Pre-specified'' denotes a plan fixed before the relevant computation on existing data. Other analyses are identified as post hoc. Independent analysis agents reimplemented major analyses from protocol text and raw events; we call this \emph{computational reproduction}. It does not add observations or replace fresh-call replication. Some post hoc reproductions had seen aggregate results. The two human auditors were authors, not AI agents.

\begin{table}[t]
\centering\small
\caption{Complementary evidence for Gemma. Each row answers a different question; results cannot be pooled as independent confirmations. CIs are 95\%.}
\label{tab:main_evidence}
\begin{tabular}{@{}>{\raggedright\arraybackslash}p{0.31\columnwidth}>{\raggedright\arraybackslash}p{0.63\columnwidth}@{}}
\toprule
Check & Observed result \\
\midrule
Fresh repetition & 34/73 selected wrong-answer candidates pass; expected-false-confirmation upper bounds 11.4--20.8. \\
Human label audit & Both readers judge 22/25 inspected helpful-message answers wrong; three split judgments. \\
Neutral text & Originally helpful wrong messages beat same-slot, equal-word neutral text by 20.0 pp [7.6, 32.2]. \\
Beyond correctness & Message-specific removal improves over the strongest tested correctness rule by 1.61 pp [1.01, 2.25]. \\
\bottomrule
\end{tabular}
\end{table}

\section{Results}
\subsection{Wrong-answer messages can help again in fresh calls}
\paragraph{Beyond one favorable replay.}
Gemma's controlled replay contains an excess over the global null of $48.3$ [24.8, 74.1] W$+$ labels and $29.1$ [9.5, 50.5] W$-$ labels. These are excess label counts, not counts of proven useful messages. About 71\% of the W$+$ excess is attributable to a pool-wide component. The message-specific excess is $14.0$ [3.0, 25.9], falling to $3.9$ [$-3.4$, 11.9] without Omni-MATH-2. Thus the aggregate alone is insufficient evidence of broadly distributed individual effects. On held-out blocks of the same run, wrong messages classified as helpful remain helpful ($+37.4$ percentage points [27.1, 47.8]); harmful messages remain harmful ($-35.5$ [$-48.1$, $-23.1$]).

\begin{figure}[t]
\centering
\includegraphics[width=\columnwidth]{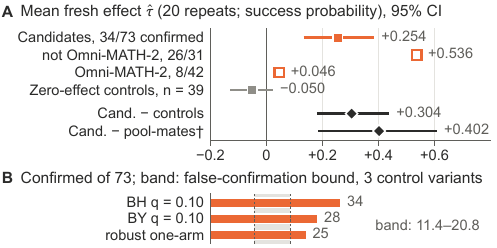}
\caption{Fresh-call evidence for selected Gemma wrong-answer messages (20 new repeats per message). Passing individual tests exceeds control-calibrated bounds on expected false confirmations; candidate-minus-control effects are positive. The conservative criterion fails when SciBench is excluded. These are selected candidates, not a prevalence sample. Protocol pre-registered; analysis computationally reproduced.}
\label{fig:t57}
\end{figure}

\paragraph{Fresh individual tests.}
The pre-registered re-test collects 20 fresh repeats for each of 73 Gemma W$+$ candidates, together with zero-original-effect controls and harmful positive controls. Under one-sided exact tests with Benjamini--Hochberg (BH) $q=0.10$, 34 candidates pass. This exceeds each control-calibrated upper bound on the \emph{expected} number of false confirmations (11.4--20.8); it does not identify a guaranteed number of true positives. The candidate-minus-control mean effect is $0.304$ [0.180, 0.438] in success probability. Gemma's own zero-effect controls pass in 2 of 39 cases, and fresh effect magnitudes average 0.35 of the original estimates, consistent with substantial selection effects.

The confirmation pattern is heterogeneous: 26 of 31 non-Omni candidates pass, compared with 8 of 42 Omni candidates (post hoc). Omni's endpoint changed between runs, so benchmark and serving conditions cannot be cleanly separated. Removing SciBench and recomputing BH yields 19 confirmations against the conservative bound of 22.7; this sensitivity prevents a claim of benchmark-wide robustness. A post hoc within-pool comparison on 17 problems nevertheless finds that fresh candidate effects exceed those of their zero-original-effect pool-mates by $0.402$ [0.185, 0.608]. Together these results establish repeatable examples and a selected-group effect, rather than a universal rate of useful errors.

\subsection{Human checks and neutral replacements support the interpretation}
\paragraph{Human evaluation.}
Two authors independently audited 13 selected problems: all 11 problems selected by the original BH screen and two illustrative masking anchors. They were blind to automated labels, selection membership, replay effects and masking outcomes. Across 49 required judgments, agreement was 42/49 (85.7\%; Cohen's $\kappa=0.61$ [0.31, 0.85]). Of the 25 W$+$ answer labels inspected on the 11 BH-selected problems, both readers judged 22 wrong and disagreed on three. No answer label met the pre-specified two-reader rule for reversal. This supports the audited examples without asserting error-free grading: reference agreement was 8/12 ($\kappa=0.14$), and eight disagreements across the audit remain unadjudicated. Upholding every single-reader answer objection still leaves a global W$+$ excess of $45.5$ [22.7, 70.6]; dropping all audited problems leaves $21.7$ [5.2, 40.5] (post hoc). The audit is selected and does not validate the entire dataset.

\paragraph{More than prompt footprint.}
In separate new calls, the originally helpful wrong-message stratum gains $25.5$ points [14.3, 36.9] over removal and $20.0$ [7.6, 32.2] over neutral text matched by slot and word count. The latter comparison shows that length and slot alone do not explain this selected-stratum effect. It does not identify which part of the content helps, and one replay per arm does not establish an effect for every individual message.

\paragraph{Answer-label and truncation sensitivity.}
A pre-specified classifier checks whether a wrong-answer message later states a reference-matching answer in its reasoning. It flags 6 of 73 W$+$ candidates. Excluding flagged messages leaves a global excess of $43.4$ [21.3, 67.5]. This is a classifier-defined check, not proof that all remaining reasoning is wrong. Treating token-cap truncations as missing gives an excess of $43.0$ [21.9, 66.8]. The latter conditions on a post-treatment property and is therefore a sensitivity analysis. Appendix C and Table~\ref{tab:human_validation} give the audit and evaluator checks.

\subsection{Correctness misses some useful distinctions between messages}
\paragraph{A held-out selection comparison.}
A message-specific rule S uses four replay blocks to choose whether to keep all five messages or remove one, then evaluates the choice on the fifth block. Rule C has the same replay budget but groups removals by the removed message's true correctness: it chooses keep-all, remove-a-wrong-message or remove-a-correct-message, uniformly within the chosen class. It therefore knows the correct labels and can estimate class-level benefits, but cannot rank messages inside a class. Lowest-index-wrong (LW) and uniformly-random-wrong (UW) rules provide additional references. C is the strongest of these three for Gemma. This is a diagnostic use of gold labels and repeated evaluation, not an inference-time algorithm.

For Gemma, S improves over C by $1.61$ points [1.01, 2.25] on held-out blocks of the same problems. The pre-specified margin survives project-wide Holm and Benjamini--Yekutieli correction. It also survives the worst pre-declared relabelling of flagged answers ($1.60$ [1.01, 2.25]), treating truncations as missing ($1.45$ [0.86, 2.10]) and restriction to programmatically decidable answers ($1.36$ [0.70, 2.11]; latter two post hoc). This supports information beyond correctness under the tested selection protocol, not a general sufficiency theorem.

\paragraph{Where the additional information lies.}
Post hoc analysis makes the result intuitive. In 256 of 262 pools containing both correct and wrong proposals, S and C have the same held-out accuracy; correctness is useful there. About 89\% of their overall margin instead comes from 367 pools in which every proposal is wrong. In those pools, the correctness label offers no ordering, yet the wrong messages can have different effects. Selecting the highest-value wrong message on training blocks yields a held-out value $4.19$ points [2.76, 5.78] above a random wrong message when Omni-MATH-2 is excluded ($6.26$ [4.68, 7.96] across all benchmarks). This measures differences in \emph{removal-defined value inside the full pool}; it does not test deploying the selected message alone. Most pools have little observed variation: 278 of 367 are at floor or ceiling across the recorded removal outcomes. Selected effects should not be mistaken for typical gains.

\paragraph{Correct answers and useful information.}
If a verified correct answer is available, this study provides no reason to replace it with a wrong one. The added lesson is that a failed answer check does not fully evaluate a message's possible contribution. The all-wrong analysis exposes this distinction particularly clearly; it does not establish that useful errors are preferable to better-generated correct solutions. On the controlled Gemma cohort, five messages versus an independent solver change accuracy by $-0.30$ points [$-2.10$, 1.60]. That baseline uses a different prompt, so the comparison supplies system context rather than isolating message availability.

\subsection{What the additional experiments do---and do not---explain}
\paragraph{Generation format changes the supply of wrong answers.}
In a pre-registered experiment on problems Gemma's independent solver answers correctly, the same role lenses produce fewer wrong proposals when asked to reason before answering. The primary comparison covers 240 problems from four non-LAB benchmarks: 52.0\% wrong under answer-first versus 32.5\% under reason-first, a difference of $-19.5$ points [$-23.3$, $-15.7$]. LAB-Bench is outside that primary and shows no detectable difference. These are real answer errors amplified by the format. The experiment measures proposal labels, not the value of the regenerated messages. Our positive value evidence is therefore scoped to answer-first Gemma pools; it does not establish that format causes useful errors.

\paragraph{Reusing content does not explain which messages help.}
In held-out Gemma traces, distinctive numeric intermediates are reused more often when the source message is present. However, the reuse lifts are similar for helpful and harmful wrong messages: $12.4$ and $12.5$ points. Across messages, reuse lift has no detectable association with value ($\bar\rho=-0.01$ [$-0.07$, 0.06]). Wrong-answer adoption differs ($-9.0$ versus $+16.8$ points), but adopting a wrong answer already entails a wrong final answer, making this association partly definitional. Intermediate matching covers only 34\% of wrong messages. These observations cannot establish a causal reasoning mechanism.

\paragraph{Effects depend on the surrounding messages.}
Pairwise message interactions exceed identical-input noise for Gemma: mean squared interaction is 0.022 [0.017, 0.029], using $K=3,4$ contrasts that preserve the integrator prompt. OSS gives 0.005 [$-0.002$, 0.013]. Thus leave-one-out value is conditional on the pool. We do not infer that two wrong answers are synergistic or that interaction signs are uniform.

\paragraph{Answer and reasoning are not simply additive.}
A separate one-block Gemma factorial experiment replaces the proposed-answer field and the remaining reasoning-related field bundle (including confidence) with neutral counterparts, individually and together. Across 301 targets from 72 questions, the interaction between these bundles is negative on Omni-MATH-2 ($-8.14$ points [$-15.57$, $-1.86$]) and inconclusive on JEEBench ($-0.54$ [$-3.24$, 2.16]). This gives no support for a positive synergy explanation and does not establish stochastic stability. It tests within-message field reveal, not pure reasoning steps (Appendix F).

\subsection{Prediction and transfer remain open}
The tested selectors do not yet turn these measurements into a practical benefit. A pre-registered learned selector evaluated on new questions gives $-0.52$ points [$-1.80$, 0.54] over random removal. On the original problems, a surface-feature ranker and a zero-shot judge yield pooled margins of $0.15$ [$-0.31$, 0.61] and $0.04$ [$-0.44$, 0.52], while reference-scored replay selection yields $2.56$ [2.09, 3.06]. Replay uses extra calls and reference answers; the comparison is not budget-matched. Tested cheap selectors also fail to distinguish useful wrong messages in the Gemma all-wrong subset (post hoc). This does not exclude process reward models, fine-tuned judges or fresh self-consistency, none of which we test. OSS has exceptions for some rules and action spaces, detailed in the appendix.

OSS also limits the central empirical claim. Its W$+$ excess is $3.6$ [$-8.7$, 18.8], and the originally helpful group does not retain a positive effect on held-out blocks. Fresh confirmation is not assessable under the original control variants; a scoring amendment gives 3 of 23, barely above its expected-false-confirmation bound, with two driven mainly by truncation. Its S$-$C margin is fragile to robustness checks. We therefore do not generalize the Gemma finding to all model families.

\section{Discussion and Limitations}
The useful distinction is between \emph{being right} and \emph{helping a particular receiver become right}. Correctness remains important, but a message includes more than its final answer. Our results motivate evaluating both properties and measuring harmful as well as helpful effects. They support using replay as a scientific diagnostic; they do not justify retaining every incorrect message or claiming an efficient pruning method.

The main limits are concrete. Selected fresh-call candidates establish repeatability rather than prevalence, with substantial benchmark and endpoint sensitivity. Held-out selection uses blocks from the same run and has not been established on fresh problems. Human validation covers a small selected set, with unadjudicated disagreements and weak reference-level agreement; remaining labels rely on automated checks. One evaluator family grades both models. Binary correctness misses changes in reasoning that do not change the final score, and finite repeats have limited power. Generation format, receiver identity and companion messages condition the effect. Within-run duplicates may also understate cross-run serving variation \citep{yuan2025numerical}.

The most informative extensions would evaluate regenerated reason-first pools with the same receiver, intervene on specific content in freshly confirmed messages, and test receiver transfer and reference-free selection on new problems. Those experiments are not results of this paper. They would determine whether the useful content can be identified prospectively and retained without the harmful content.

\section{Conclusion}
A wrong answer does not necessarily make its message useless. In the tested Gemma setting, some wrong-answer messages repeatedly improve a receiver's chance of success, while others harm it. Fresh calls, human label checks and neutral replacements strengthen this finding; benchmark sensitivity, format dependence and weaker OSS evidence define its limits. Answer correctness and trajectory value should therefore be measured separately. Identifying useful errors before observing the receiver's outcome remains an open practical problem.

\bibliography{references_v2}
\clearpage
\appendix
\raggedbottom
\section*{Appendix: guide to the evidence}
Appendix A gives the protocol, decoding settings and information boundaries; B defines the replay estimators and nulls. C reports human label validation and D the fresh-call and neutral-text tests. E details correctness-based selection and its population; F separates text correspondence, between-message interaction and within-message intervention. G gives the format experiment, H the predictor comparisons and I the robustness checks. J expands the real example from the introduction. K records computational provenance and L provides the core prompt text. Repeated analyses of the same events are supporting checks, not additional data collections.

\section{A. Protocol, decoding and information boundaries}
Role-conditioned drafting generates five complementary attempts without instructing any sender to make mistakes. Recruiter and hypothesizer prompts explicitly request good-faith solutions. The hypothesis schema orders fields as role, proposed final answer, reasoning direction, key idea, critical assumption, what another solver should check, and confidence. The integrator is told that proposals can be wrong and that it should verify and synthesize its own answer. This instruction is part of the experimental setting: our evidence does not establish the same effects for an integrator instructed simply to vote.

The evaluator receives the problem, final answer and evaluator-only reference material. Neither recruiter, hypothesizer nor integrator receives the gold answer, correctness label or measured trajectory value during ordinary generation. The offline selection diagnostics use gold labels and replay outcomes explicitly; they do not retroactively become information available to the original integrator. Core prompt text is in Appendix L; benchmark guidance supplies answer-format instructions.

\begin{table}[ht]
\centering\small
\caption{Configured original-protocol decoding settings, shared across families. Token limits are completion caps.}
\label{tab:decoding}
\begin{tabular}{lrr}\toprule
Role & Temperature & Token cap \\
\midrule
Recruiter & 0 & 2,048 \\
Each hypothesizer & 0.7 & 4,096 \\
Integrator / independent solver & 0 & 4,096 \\
Evaluator (gpt-oss-120b) & 0 & 4,096 \\
\bottomrule\end{tabular}
\end{table}
No reasoning-effort parameter was sent. The provider's effective default and any Gemma thinking mode were not recorded, so an effort level cannot be inferred. Per-message completion tokens and the effective cap sent on every request were also not recorded. Truncation checks therefore use the available output and tokenizer proxies. The science-task OSS independent-solver baseline uses a corrected 4,096-token cap.

\paragraph{The independent-solver comparison.}
$K=0$ invokes a separate prompt asking the same model to solve independently. It is not the integrator prompt with an empty message block. A $K=1$ versus $K=0$ contrast changes both prompt and message availability; the original archive's $K=2$--5 single-removal contrasts preserve the integrator template. Our controlled five-message replay and primary $K=3,4$ interaction estimates avoid the $K=0$ contrast. Complete-system accuracy is therefore context, not an equal-budget causal test of messaging.

\section{B. Replay estimators, uncertainty and nulls}
For problem $x$, message $i$ and block $b$, let $Y^F_{xb1},Y^F_{xb2}$ be full-pool outcomes and $Y^{-i}_{xb}$ the outcome after removing $i$. The block difference and its mean are
\begin{align}
d_{xib}&=\tfrac12(Y^F_{xb1}+Y^F_{xb2})-Y^{-i}_{xb},\\
\hat\tau_{xi}&=\tfrac{1}{B}\sum_{b=1}^{B}d_{xib}.
\end{align}
The original controlled analysis uses $B=5$. Full-pool calls are shared by all removals within a block, inducing dependence across a problem's messages. Bootstrap resampling therefore clusters on problems; it does not treat every replay call as an independent example. Stable helpful and harmful labels require the bootstrap 95\% interval of the mean block difference to lie wholly above or below zero, respectively. Intervals crossing zero leave the sign unresolved. W$\pm$ and C$\pm$ cross these labels with proposal correctness.

For the global null we enumerate arrangements of a block's seven binary call outcomes over the two full and five removal slots, with their multiplicities. For the removal-only null we hold both full outcomes fixed and enumerate arrangements over the five removal slots. Applying the unchanged labelling rule gives the expected label count under each null; the reported excess is the observed count minus that expectation. These are exchangeability-based references. They do not correct arbitrary time-dependent serving changes, motivating randomized fresh calls and controls.

Let $N_{\rm obs}$ be an observed count and $N_G,N_R$ the expected global and removal-only counts. The decomposition
\begin{equation}
N_{\rm obs}-N_G=(N_R-N_G)+(N_{\rm obs}-N_R)
\end{equation}
is an identity: the first component is compatible with a shared pool-wide full-versus-removal advantage, while the second distinguishes among removals. A global excess need not mean every labelled message is special. Table~\ref{tab:bracket} reports both directions of wrong-message effects and the benchmark sensitivity.

\paragraph{Why helpful shares among flips can mislead.}
Write $p_+=\Pr(Y^F=1,Y^R=0)$ and $p_-=\Pr(Y^F=0,Y^R=1)$. Then $\tau=p_+-p_-$, the flip rate is $f=p_++p_-$, and the helpful share among flips is
\begin{equation}
\frac{p_+}{p_++p_-}=\frac12+\frac{\tau}{2f},\qquad f>0.
\end{equation}
The pooled analogue depends on the mean effect, which can be small when help and harm offset. Gemma's observed share (55.0\%) does not exceed its matched identical-input null (56.9\%); OSS is below its null (42.8\% versus 50.9\%). We do not use those shares as evidence of useful errors. Signed effects and null-calibrated counts answer a different question: whether changes persist in either direction.
\begin{table*}[t]
\centering\small
\caption{Global excess and message-specific excess of repeatable labels. The global permutation null varies all seven call slots; the removal-only null fixes full-pool outcomes. The global excess equals a pool-wide component plus the removal-only excess before rounding. Decomposition is post hoc and computationally reproduced; 95\% problem-bootstrap CIs.}
\label{tab:bracket}
\begin{tabular}{@{}>{\raggedright\arraybackslash}p{0.42\textwidth}>{\raggedright\arraybackslash}p{0.25\textwidth}>{\raggedright\arraybackslash}p{0.27\textwidth}@{}}
\toprule
Family and label & Global excess & Message-specific excess \\
\midrule
Gemma W$+$ & $48.3$ [24.8, 74.1] & $14.0$ [3.0, 25.9] \\ %
Gemma W$-$ & $29.1$ [9.5, 50.5] & $18.8$ [9.5, 28.8] \\ %
Gemma C$+$ & $25.9$ [14.8, 38.2] & $22.6$ [13.1, 33.1] \\ %
Gemma C$-$ & $2.3$ [$-7.5$, 16.2] & --- \\ %
OSS W$+$ & $3.6$ [$-8.7$, 18.8] & --- \\ %
OSS W$-$ & $20.1$ [8.1, 33.3] & --- \\ %
OSS C$+$ & $39.4$ [21.9, 57.8] & --- \\ %
\midrule
\multicolumn{3}{l}{\emph{Gemma W$+$ sensitivities}} \\
Exclude Omni-MATH-2 & $15.8$ [2.1, 31.4] & $3.9$ [$-3.4$, 11.9] \\ %
Exclude classifier-defined latent-correct messages & $43.4$ [21.3, 67.5] & $11.9$ [1.2, 23.2] \\ %
Treat truncations as missing & $43.0$ [21.9, 66.8] & --- \\ %
Drop all 13 human-audited problems & $21.7$ [5.2, 40.5] & --- \\ %

\bottomrule
\end{tabular}
\par\smallskip{\footnotesize The Gemma W$+$ pool-wide component is $34.2$ [13.4, 57.0], about 71\% of its global excess. Thus the global count alone cannot establish distinct effects for individual messages. For OSS, the C$+$ excess falls to $12.7$ [4.6, 21.7] when truncations are treated as missing; it is partly an output-reliability effect. Missing cells are not estimated here, rather than zero. %
}
\end{table*}

\section{C. Human evaluation and answer-label validity}
Table~\ref{tab:human_validation} gives aggregate judgments only. The human audit evaluates reference and answer correctness; it does not ask people whether a message helps a model. Readers independently saw problem and answer material while being blinded to automatic correctness, parser flags, original BH selection, replay effects and masking outcomes. The selected set includes all 11 problems from the original BH screen and two masking anchors. Requiring both readers to support a reversal prevents a single objection from silently changing the analysis, but leaves disagreements unresolved. Sensitivity analyses show what happens if single-reader answer objections are upheld instead.

\paragraph{Automatic checks outside the human sample.}
On programmatically decidable Gemma items (option letters or integers), a programmatic verifier agrees with the LLM evaluator on 99.96--100\% of labels (post hoc). This narrower subset does not establish correctness on symbolic, conditional or ambiguous-reference items. A separate screen detects messages whose reasoning contains a reference-matching answer despite a wrong committed answer. Its latent-correct flags have 68.5\% precision on an AI-read validation set. A negative flag therefore does not validate the reasoning or rule out every hidden correct answer. The original scoring convention labels the committed answer, not every statement in the text.

\paragraph{Evaluator dependence.}
Two additional evaluator families retain 0.66 [0.52, 0.78] of the original Gemma W$+$ labels, compared with 0.76 under re-judging by the original evaluator. For OSS these rates are 0.42 [0.29, 0.56] and 0.74. These are conditional retention rates in a balanced audit, not population correctness estimates. They show meaningful evaluator sensitivity even where checks support the headline direction.
\begin{table*}[t]
\centering\small
\caption{Blinded human validation of selected Gemma answer labels. Two authors independently read 13 problems: all 11 selected by the original BH screen and two masking anchors. This evaluates answer correctness, not human judgments of message usefulness. Intervals are 95\%.}
\label{tab:human_validation}
\begin{tabular}{@{}>{\raggedright\arraybackslash}p{0.57\textwidth}>{\raggedright\arraybackslash}p{0.40\textwidth}@{}}
\toprule
Quantity & Result \\
\midrule
Required judgments; overall agreement & 49; 42/49 (85.7\%) \\ %
Cohen's $\kappa$ (three categories) & 0.61 [0.31, 0.85] \\ %
Agreement on answers & 34/37 (91.9\%) \\ %
Agreement on reference answers & 8/12 (66.7\%); $\kappa=0.14$ \\ %
W$+$ labels judged wrong by both readers & 22/25; three split judgments \\ %
All wrong-labelled messages on the 11 selected problems & 48/54 judged wrong by both; six split \\ %
Labels reversed by the pre-specified two-reader rule & 0 \\ %
Single-reader disagreements across the audit & 8; no human adjudication performed \\ %
\midrule
\multicolumn{2}{l}{\emph{Global W$+$ excess under post hoc sensitivity analyses (label counts)}} \\
Original answer labels & $48.3$ [24.8, 74.1] \\ %
Uphold every single-reader answer objection & $45.5$ [22.7, 70.6] \\ %
Drop every problem with any disagreement & $36.2$ [15.7, 59.5] \\ %
Drop all 13 audited problems & $21.7$ [5.2, 40.5] \\ %

\bottomrule
\end{tabular}
\par\smallskip{\footnotesize No label change means that the consensus rule was not triggered; it does not mean that all readers agreed. An automated equivalence screen assessed the 48 W$+$ labels outside the original 11 BH-selected problems; 15 were undecidable. In the extreme sensitivity treating all 48 as correct, the W$+$ excess becomes $3.1$ [$-12.4$, 22.3]. This is a dependency on label quality, not an estimated error rate. %
}
\end{table*}

\section{D. Fresh-call re-test and neutral replacement}
The individual fresh re-test fixes candidates using the original replay, then collects 20 new repeated comparisons. Each target receives full-pool and target-removal calls. Wrong messages with zero original effect supply negative controls; originally harmful messages supply positive controls. Individual one-sided exact within-block permutation tests are corrected with BH at $q=0.10$. The false-confirmation calibration applies the identical decision rule to controls and converts the uncertainty in their pass rate to an upper bound on the expected number of candidate false confirmations. It is not a confidence bound on the realized number of false discoveries.

The three readings vary the OSS control scoring and thus the pooled control calibration. Gemma passes all three, with 34/73 versus expected-false-confirmation bounds 20.8, 14.3 and 11.4. OSS's original variants cannot meet the evaluability criterion (0/23 and 11/23 evaluable). Its documented amended scoring yields 3/23 confirmations versus 2.89; two are mainly truncation-type, and the candidate-minus-control interval includes zero. The amended result is not evidence of a stable general OSS wrong-helpful effect.

Post hoc benchmark removal recomputes the full decision procedure, including BH; subtracting a benchmark's original confirmations is not equivalent. Without SciBench the Gemma count is 19 versus the conservative bound 22.7. Outside Omni, 26/31 pass; Omni has 8/42 and an endpoint change. These selected-subset results are not domain prevalence estimates. In the fresh within-pool comparison, candidate-minus-control effects are positive in 13 of 16 nonzero problems out of 17 total, sign-test $p=0.011$.

\paragraph{Neutral replacement and missingness.}
For Gemma, fresh full/removal/neutral calls cover 969 of 1,000 original problems, with one call per arm. Each target is replaced in its original slot by neutral redaction text with equal word count. Public-tokenizer counts are approximately matched; they do not establish exact endpoint-token identity. The surviving share $(F-N)/(F-R)$ is 0.85 [0.31, 1.60] over all messages and 0.79 [0.40, 1.12] for originally helpful wrong messages. A ratio interval extending above one is possible; it is not a probability. The neutral effect includes a detectable footprint for correct-answer messages. The 31 missing problems are not assumed missing at random, and OSS coverage is incomplete, so the strongest inference is for the completed Gemma cohort.
\begin{table*}[t]
\centering\small
\caption{Repeatability of selected message effects. Fresh re-test: 20 new repeats per target, exact one-sided tests and BH $q=0.10$. Bounds concern expected false confirmations under control-calibrated nulls. Held-out and neutral-replacement strata are selected on original replay data; their denominators differ from the individual re-test. Intervals are 95\%.}
\label{tab:repeatability}
\begin{tabular}{@{}>{\raggedright\arraybackslash}p{0.42\textwidth}>{\raggedright\arraybackslash}p{0.25\textwidth}>{\raggedright\arraybackslash}p{0.27\textwidth}@{}}
\toprule
Comparison & Gemma & OSS \\
\midrule
\multicolumn{3}{l}{\emph{Individual fresh re-test (pre-registered; computationally reproduced)}} \\
Confirmed candidates / tested & 34/73 & Original protocol not assessable \\ %
Control variants 1 / 2 / 3: upper bound on expected false confirmations & 20.8 / 14.3 / 11.4 & Amended variant 3: 2.89 \\ %
Candidate-minus-control effect (success probability) & $0.304$ [0.180, 0.438] & Amended: $0.094$ [$-0.012$, 0.217] \\ %
BY / robust one-arm confirmations & 28 / 25 & --- \\ %
Own zero-original-effect controls passing & 2/39 & --- \\ %
Harmful positive controls detected & 5/8 & --- \\ %
Non-Omni / Omni candidates passing (post hoc) & 26/31 / 8/42 & --- \\ %
Exclude SciBench, recompute BH and bound & 19 versus 22.7 & --- \\ %
Within-pool candidate $-$ control (17 problems; post hoc) & $0.402$ [0.185, 0.608] & --- \\ %
\midrule
\multicolumn{3}{l}{\emph{Select on blocks 0--1, evaluate on blocks 2--4 (percentage points)}} \\
Originally helpful wrong messages & $37.4$ [27.1, 47.8]; 153 messages, 62 problems & $0.0$ [$-7.0$, 7.2]; 124 messages, 64 problems \\ %
Originally harmful wrong messages & $-35.5$ [$-48.1$, $-23.1$]; 128 messages, 57 problems & $-7.9$ [$-13.6$, $-2.1$]; 158 messages, 100 problems \\ %
\midrule
\multicolumn{3}{l}{\emph{Fresh neutral-text control, Gemma (pp; one replay per arm)}} \\
Selected stratum; slots / problems & Full $-$ removal & Full $-$ neutral text \\
Helpful wrong; 165 / 69 & $25.5$ [14.3, 36.9] & $20.0$ [7.6, 32.2] \\ %
Harmful wrong; 151 / 65 & $-12.3$ [$-22.7$, $-2.0$] & $-8.3$ [$-17.3$, 0.4] \\ %
Zero-effect wrong; 2,135 / 563 & $-0.1$ [$-0.8$, 0.5] & $0.3$ [$-0.4$, 1.0] \\ %

\bottomrule
\end{tabular}
\par\smallskip{\footnotesize Gemma bounds pool zero-effect controls across families, hence their dependence on the OSS control variant. The original OSS variants have only 0/23 and 11/23 evaluable candidates; the amended variant gives 3/23 confirmations, with two mainly truncation-related. It is insufficient evidence for a robust OSS effect. Gemma fresh/original effect ratio is 0.35; per-message power is 0.22 at $\tau=0.3$ and 0.87 at 0.5. Neutral replacement matches slot and word count, not exact endpoint-token count. %
}
\end{table*}

\section{E. Selection rules, pool composition and costs}
For each held-out block, S estimates each of the five single-removal actions on the other four blocks and compares them with keep-all. C averages removal outcomes within the wrong and correct proposal classes, picks the highest-valued class or keep-all on training blocks, and randomizes uniformly within a chosen removal class. Its true correctness labels and replay budget make it a strong diagnostic comparator, not an implementable verifier. LW removes the lowest-index wrong message and UW removes a uniformly selected wrong message when one is available; their keep-all fallback applies otherwise. S breaks ties in favor of keep-all, then the lowest removal index. The reconstructed C rule favors keep-all, then the wrong-answer class before the correct-answer class. S$-$C compares the chosen actions on the same held-out blocks, with paired problem bootstrap.

The controlled cohort has 1,000 problems per family, but the reconstructed selection comparison has 999 eligible Gemma pools. This denominator is separate from the 969-problem neutral control and the 73-candidate re-test. Within each valid five-block pool, the training/evaluation split rotates across blocks. It holds out outcomes for the same problems and messages; it does not test transfer to new tasks or a new serving run.

\paragraph{Selecting a high-value wrong message.}
For all-wrong pools, let $i_b^*$ be chosen uniformly among the messages maximizing training-block $\hat\tau_i$. Define
\begin{equation}
U=100\,\mathbb E_{x,b,i_b^*}\left[d_{xi_b^*b}-\frac{1}{5}\sum_{i=1}^5 d_{xib}\right].
\end{equation}
The expectation includes uniform averaging across tied maximizers, unlike S's lowest-index tie break. This contrasts removal-defined value with the pool's average; its units are percentage points of integrator success probability. It does not compare two keep-one policies. The analysis was motivated after inspecting the original data and computationally reproduced, partly blind to earlier aggregate estimates.

Of 367 Gemma all-wrong pools, 278 are saturated in the observed removal outcomes: 111 at the floor and 167 at the ceiling. The other 89 account for the measured variation. Seventy-two pools have positive held-out $U$, versus 37 under the chance reference. These outcome-derived strata describe concentration; conditioning on them does not estimate the effect in a prospectively identifiable population. The full-cohort and Omni-excluded estimates in Table~\ref{tab:sufficiency} keep saturated pools in the denominator.

\paragraph{Variance explained by correctness.}
We center each message's value within its pool, $\taut_i=\tau_i-\frac15\sum_j\tau_j$, and compare reliable within-pool variance with the variance explained by correctness classes. Split-block covariance corrects within-run measurement noise. Correctness cannot explain within-pool differences when all labels agree, so the overall share depends on the mixture of pool types. The Gemma overall share 0.24 [0.14, 0.33] and mixed-pool share 0.75 must be read together. Neither is an AUROC for whether a message truly helps versus harms.

\paragraph{Budget and practical meaning.}
One training block uses seven integrator calls; grading all seven adds seven evaluator calls. Four training blocks therefore cost 56 calls per problem, excluding pool generation and held-out assessment. Gold references or an adequate verifier are needed to score those calls. A post hoc reduction to one training block retains 0.700 [0.515, 0.885] of Gemma's S$-$C margin (0.546 without Omni), while C retains 0.858 of its gain. This same-run diagnostic was computationally reproduced; it is not a reference-free deployment claim, and OSS does not show the same result. Table~\ref{tab:baserates} gives system-level context without implying cost equivalence.

\begin{figure}[t]
\centering
\includegraphics[width=\columnwidth]{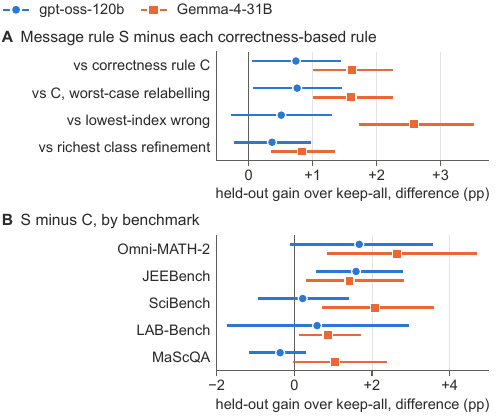}
\caption{Held-out accuracy margins for message-specific selection. (A) Comparison with correctness rules and refinements; (B) S$-$C by benchmark. Paired problem-bootstrap 95\% intervals. These are same-run, same-problem diagnostics; Table~\ref{tab:sufficiency} gives the population and sensitivity definitions.}
\label{fig:sufficiency}
\end{figure}

\begin{table*}[t]
\centering\small
\caption{Message-specific selection versus correctness-based selection. Both rules use four training blocks of the same problems and are evaluated on the fifth. S may keep all messages or remove one; C selects a correctness class and removes a uniformly random member, or keeps all. Percentage-point differences and paired problem-bootstrap 95\% CIs; 1,000 OSS and 999 eligible Gemma pools. The all-wrong analysis is post hoc.}
\label{tab:sufficiency}
\begin{tabular}{@{}>{\raggedright\arraybackslash}p{0.42\textwidth}>{\raggedright\arraybackslash}p{0.25\textwidth}>{\raggedright\arraybackslash}p{0.27\textwidth}@{}}
\toprule
Comparison & OSS & Gemma \\
\midrule
S $-$ C (pre-specified primary) & $0.74$ [0.05, 1.44] & $1.61$ [1.01, 2.25] \\ %
S $-$ lowest-index-wrong removal & $0.51$ [$-0.28$, 1.30] & $2.58$ [1.73, 3.52] \\ %
S $-$ C, worst pre-declared label perturbation & $0.76$ [0.07, 1.46] & $1.60$ [1.01, 2.25] \\ %
S $-$ C, truncations treated as missing & $0.44$ [$-0.07$, 0.97] & $1.45$ [0.86, 2.10] \\ %
S $-$ C, decidable answers only & --- & $1.36$ [0.70, 2.11] \\ %
Problems to remove before CI reaches zero & 1 & 23 \\ %
Project-wide Holm/BY conclusion & Does not survive & Survives \\ %
\midrule
\multicolumn{3}{l}{\emph{Localization and value ranking (post hoc)}} \\
Share of S $-$ C margin from all-wrong pools & --- & 0.89 [0.77, 0.97] \\ %
$U$: selected $-$ random wrong message value, all-wrong pools & --- & $6.26$ [4.68, 7.96] \\ %
$U$, excluding Omni-MATH-2 & --- & $4.19$ [2.76, 5.78] \\ %
Correctness-explained share of reliable within-pool variance & 0.59 [0.44, 0.76] & 0.24 [0.14, 0.33] \\ %
Same share in mixed-correctness pools & 0.70 & 0.75 \\ %
S $-$ refined correctness classes & $0.37$ [$-0.24$, 0.98] & $0.84$ [0.36, 1.35] \\ %
S $-$ equally fine random class refinement & $0.51$ [$-0.07$, 1.11] & $0.84$ [0.42, 1.28] \\ %

\bottomrule
\end{tabular}
\par\smallskip{\footnotesize $U$ selects the highest removal-defined value on training blocks and scores it on the held-out block; it does not measure keeping that message alone. The variance share uses within-pool centering and a split-block reliability correction; all-wrong pools have zero correctness-explained variance by construction. Refined classes combine correctness, corroboration, confidence, revision flags and reference-based answer closeness; the latter is an oracle feature. Matching performance of random refinement prevents attributing its gain to these semantics. All quoted post hoc values were computationally reproduced; some reproductions had seen aggregate results. Truncation analysis retains 757 OSS and 968 Gemma problems. %
}
\end{table*}

\begin{table*}[t]
\centering\small
\caption{System-level context and pool composition (post hoc, computationally reproduced). Accuracy differences are percentage points. The independent solver has its own prompt, so five-message-versus-solo contrasts do not isolate the effect of message availability.}
\label{tab:baserates}
\begin{tabular}{@{}>{\raggedright\arraybackslash}p{0.42\textwidth}>{\raggedright\arraybackslash}p{0.25\textwidth}>{\raggedright\arraybackslash}p{0.27\textwidth}@{}}
\toprule
Quantity & Gemma & OSS \\
\midrule
Five messages $-$ independent solver, controlled cohort & $-0.30$ [$-2.10$, 1.60] & --- \\ %
Five messages $-$ independent solver, full archive & $0.34$ & $1.67$ [0.60, 2.74] \\ %
Replay keep-or-remove-one $-$ keep-all, held-out block & $2.61$ & $1.68$ \\ %
Mean effect of a wrong message & $0.43$ & $-1.66$ \\ %
Mixed / all-wrong pools (999 eligible Gemma pools) & 262 / 367 & --- \\ %
All-wrong pools whose problem the solo solver answers correctly & 65\% & --- \\ %

\bottomrule
\end{tabular}
\par\smallskip{\footnotesize Positive and negative wrong-message effects can cancel in the mean. The solo-solved fraction concerns a different prompt and does not explain which individual messages help. Point-only summaries are reported descriptively; no significance is inferred from them.}
\end{table*}

\section{F. Context dependence and content interventions}
\paragraph{Between-message interactions.}
For a pool and two messages $i,j$, the interaction contrast is
\begin{equation}
I_{ij}=Y(P)-Y(P\setminus i)-Y(P\setminus j)+Y(P\setminus\{i,j\}).
\end{equation}
Mean squared interaction, corrected using identical-input noise, tests whether non-additivity exceeds stochastic variation. The primary uses $K=3,4$ so every cell retains the integrator template. It detects context dependence for Gemma but does not assign an individual message a stable interaction sign. $K=2$ delete-both contrasts invoke the separate independent-solver prompt and cannot support a semantic claim about pairs of wrong answers.

\paragraph{Text correspondence.}
Distinctive numeric or mathematical intermediates are matched between a message and the integrator output with versus without that message, with placebo checks. This covers 34\% of wrong messages and excludes general semantic reasoning without such a match. Wrong-answer adoption uses answer matching. Two-fold cross-fitting separates the data used for value and correspondence estimates. Similar reuse for helpful and harmful messages means that observing reused content alone does not explain a benefit. Adoption is partly tied to the outcome definition. These checks do not establish mediation or faithful reasoning traces.
\begin{table*}[t]
\centering\small
\caption{Context dependence and text correspondence. Pairwise mean squared interaction (MSI) uses $K=3,4$ templates and subtracts identical-input noise. Reuse and adoption are held-out Gemma percentage-point lifts with the target present versus removed. Numeric-intermediate matching covers 34\% of wrong messages; 95\% CIs.}
\label{tab:structure}
\begin{tabular}{@{}>{\raggedright\arraybackslash}p{0.42\textwidth}>{\raggedright\arraybackslash}p{0.25\textwidth}>{\raggedright\arraybackslash}p{0.27\textwidth}@{}}
\toprule
Quantity & OSS & Gemma \\
\midrule
MSI beyond noise & 0.005 [$-0.002$, 0.013] & 0.022 [0.017, 0.029] \\ %
\midrule
Gemma wrong-message stratum & Intermediate reuse lift & Wrong-answer adoption lift \\
Helpful & $12.4$ [4.4, 21.3] & $-9.0$ [$-15.8$, $-2.9$] \\ %
Harmful & $12.5$ [2.5, 21.2] & $16.8$ [9.8, 23.9] \\ %
Cross-fitted correlation with value & $-0.01$ [$-0.07$, 0.06] & $-0.28$ [$-0.35$, $-0.19$] \\ %

\bottomrule
\end{tabular}
\par\smallskip{\footnotesize Interaction magnitude does not determine its sign or mechanism. We exclude $K=2$ semantic interpretations because deleting both messages invokes the separate solo prompt. Reuse is not causal mediation; wrong-answer adoption is mechanically coupled to final incorrectness. %
}
\end{table*}

\paragraph{Within-message field intervention.}
A separate factorial experiment holds four peer messages and the target slot fixed. $A$ is the proposed-answer field. $R$ is the bundle of reasoning direction, key idea, critical assumption, what to check, and confidence; it is not pure chain-of-thought text. A 1 retains the original field or bundle and a 0 supplies deterministic neutral text. Local-tokenizer matching applies to edited fields, with whole rendered Gemma blocks within two tokens, rather than an exact guarantee about endpoint tokenization. The outcome interaction is
\begin{equation}
\theta=\mathbb E[Y_{R1A1}-Y_{R1A0}-Y_{R0A1}+Y_{R0A0}].
\end{equation}
Targets are averaged within questions and then questions receive equal weight. The completed Gemma experiment is one four-arm block: 1,204 valid outcomes for 301 targets across 72 questions. Question-cluster bootstrap uses 5,000 resamples. Omni has 155 targets/35 questions and $\theta=-8.14$ pp [$-15.57$, $-1.86$]; JEE has 146/37 and $-0.54$ [$-3.24$, 2.16]. These are sub-additive or inconclusive interactions, not positive synergy. The design is mostly saturated (28/35 Omni and 33/37 JEE question clusters), which limits precision. The single block does not estimate stochastic stability; enlarged planned panels are not treated as completed data.

\section{G. Generation-format experiment}
The pre-registered comparison regenerates messages for the same problems and role lenses under answer-first and reason-first prompts at temperature 0.7. Eligibility requires the Gemma independent solver to answer the problem correctly; the results do not describe arbitrary unsolved tasks. Sixty problems per benchmark are evaluated. The primary contrast pools four non-LAB benchmarks (240 problems); LAB is a separate control. Table~\ref{tab:format} reports proposal-error rates and intervals. A confidence interval excluding zero is not automatically a pass under the protocol's minimum-effect criterion.

Reason-first prompting reduces answer errors, particularly on SciBench. It also changes the share of wrong proposals that later contain a reference-matching answer. Neither observation identifies the value of a newly generated message. A value comparison would require replaying the regenerated pools; simply relabelling old messages changes correctness-based statistics even if the receiver's behavior is unchanged. We therefore do not use a reduction of the correctness-selection gap as evidence that format explains wrong-but-useful value.
\begin{table*}[t]
\centering
\small
\caption{Generation-format effects on wrong-proposal rates, not receiver value. Pre-registered format
experiment (Gemma, $T{=}0.7$, problems Gemma's own independent solver answers correctly; same problems and role
lenses in both arms): wrong-proposal rate when the hypothesizer states its answer before (answer-first, AF) or after
(reason-first, RF) its reasoning. The analysis was computationally reproduced.}
\label{tab:format}
\begin{adjustbox}{max width=\textwidth}
\begin{tabular}{lrrr}
\toprule
\multicolumn{4}{l}{\textbf{Proposal-error rates} \hfill wrong proposals (\%), 60 problems per benchmark} \\
 & AF & RF & RF $-$ AF (pp) [95\% CI] \\
\midrule
SciBench      & 61.8 & 27.5 & $-34.3\;[-42.2, -26.4]$ \\   %
JEEBench      & 58.9 & 41.0 & $-17.9\;[-25.6, -10.2]$ \\   %
MaScQA        & 32.3 & 15.0 & $-17.3\;[-25.7, -9.7]$ \\    %
Omni-MATH-2   & 55.0 & 46.3 & $-8.7\;[-14.0, -3.7]$ \\     %
LAB-Bench (outside primary) & 12.7 & 11.0 & $-1.7\;[-5.0, 1.7]$ \\ %
\midrule
Primary: 4 non-LAB benchmarks, 240 problems & 52.0 & 32.5 & $-19.5\;[-23.3, -15.7]$ \\ %
\quad strict programmatically decidable subset ($n{=}100$) & & & $-8.8\;[-13.1, -4.8]$ (inconclusive) \\ %
\quad AF reproduces the original wrong rate & 52.0 vs 51.4 & & \\                %
Wrong messages whose own reasoning reaches gold (latent-correct) & 9.6 & 1.7 & \\        %
\bottomrule
\end{tabular}
\end{adjustbox}\par\smallskip
{\footnotesize The primary includes four non-LAB benchmarks (240 problems); LAB is separate. A confidence interval excluding zero is not the same as passing the pre-specified substantive-effect criterion: Omni and the decidable subset are inconclusive by that criterion. The manipulation measures labels, not receiver value. Latent-correct flags are automated, with 68.5\% precision on an AI-read validation set. Computational reproduction is independent code on the same data.}
\end{table*}

\section{H. Predictors and action spaces}
Table~\ref{tab:boundary} compares forced removal against random removal, separately from the keep-or-remove-one action space of S and C. The ranker uses pre-integration features with leave-one-benchmark-out evaluation. The zero-shot judge sees the integrator's input and predicts which removal is least harmful; on OSS pools it is the same model family as the integrator. The pooled nulls do not mean every predictor is useless on every task: an OSS lens prior gains 0.51 [0.03, 0.99] points before multiplicity correction, and consensus/confidence have OSS exceptions in the keep-one action space. Eleven tested cheap selectors do not beat random selection inside Gemma all-wrong pools in post hoc analysis. Those methods are a bounded set, not all possible predictors.

The separate learned-selector experiment trains on 100 problems and uses 100 locked MGSM/logic-grid test questions; its $-0.52$ [$-1.80$, 0.54] margin does not improve on random removal. We do not retune that experiment on its test set. Process reward models \citep{lightman2023verify,zhang2025lessonsprm}, step-error benchmarks \citep{zheng2025processbench,song2025prmbench}, fine-tuned judges and fresh self-consistency suggest additional candidates, but they were not evaluated here. A valid practical comparison would account for their extra calls, verifier access and action space.
\begin{table*}[t]
\centering\small
\caption{Tested removal rules on the 2,000 controlled pools, pooled across families. Each rule must remove exactly one message; the reference is uniform random removal. Gains are percentage points with problem-cluster 95\% CIs. AUC compares pairs within a pool whose removal outcomes differ. Learned ranker uses leave-one-benchmark-out evaluation.}
\label{tab:boundary}
\begin{tabular}{@{}>{\raggedright\arraybackslash}p{0.42\textwidth}>{\raggedright\arraybackslash}p{0.25\textwidth}>{\raggedright\arraybackslash}p{0.27\textwidth}@{}}
\toprule
Rule & Gain over random & Within-pool AUC \\
\midrule
Replay: choose on four blocks, score fifth & $2.56$ [2.09, 3.06] & --- \\ %
Gold proposal correctness & $0.93$ [0.69, 1.17] & 0.587 \\ %
Learned ranker: 25 pre-integration features & $0.15$ [$-0.31$, 0.61] & 0.506 \\ %
Zero-shot LLM judge & $0.04$ [$-0.44$, 0.52] & 0.508 \\ %
Answer consensus & $-0.03$ [$-0.38$, 0.30] & 0.508 \\ %
Self-reported confidence & $0.06$ [$-0.17$, 0.29] & 0.502 \\ %
Message length & $0.11$ [$-0.38$, 0.61] & 0.487 \\ %
Random removal & 0 & 0.500 \\

\bottomrule
\end{tabular}
\par\smallskip{\footnotesize The ranker and judge use the same pools and evaluation harness, so these are not independent experiments. Planted-signal controls recover 96\% and 96.5\% of the in-sample maximum. Replay requires reference-scored calls and is not cost-matched to these predictors. On a separate set of 100 locked MGSM/logic-grid test questions, a pre-registered learned selector gives $-0.52$ [$-1.80$, 0.54]. %
}
\end{table*}

\section{I. Robustness, multiplicity and interpretation}
The robustness battery fixes perturbation rules before computation: leave-one-problem-out, deletion of the most favorable problems, leave-one-benchmark-out and benchmark-stratified bootstrap. It separates sensitivity to a small set of examples from benchmark concentration. Table~\ref{tab:robustness} summarizes central contrasts; Table~\ref{tab:multiplicity} gives project-wide correction across 11 specified headline tests. Post hoc localization is not promoted to a pre-registered primary by computational reproduction.

Truncation is substantially more common for OSS (15.5\% of events) than Gemma (0.67\%). Removal changes the prompt and may change whether an output reaches its cap; conditioning on complete outputs is post-treatment selection. Consequently, truncation-as-missing checks bound interpretation rather than isolate a pure answer-content effect. The fresh re-test also exhibits substantial effect shrinkage and endpoint sensitivity. A claim can pass one check and fail another: global excess, relative message ranking and fresh individual confirmation measure different quantities.

Two exploratory routes do not support stronger claims. A conjunction of stable labels and superiority to pool-mates failed its frozen synthetic-bias criterion and is not used as confirmation. A numerical near-miss metric failed two validation checks and did not cover 27 of 73 helpful candidates; it cannot rule out all forms of near-correct content. These limitations motivate reporting direct content controls and fresh re-tests instead of a complete causal explanation.
\begin{table*}[t]
\centering\small
\caption{Robustness of central claims under rules fixed before perturbation. Selection margins are percentage points; excesses are label counts. The fresh test compares observed confirmations with the conservative upper bound on expected false confirmations after each exclusion. Intervals are 95\%.}
\label{tab:robustness}
\begin{tabular}{@{}>{\raggedright\arraybackslash}p{0.42\textwidth}>{\raggedright\arraybackslash}p{0.25\textwidth}>{\raggedright\arraybackslash}p{0.27\textwidth}@{}}
\toprule
Claim & Sensitivity & Result \\
\midrule
Gemma S $-$ C & Five most favorable problems removed & $1.21$ [0.72, 1.74] \\ %
Gemma S $-$ C & Worst benchmark exclusion: Omni & CI [0.77, 1.99] \\ %
OSS S $-$ C & One most favorable problem removed & CI [$-0.01$, 1.32] \\ %
Gemma global W$+$ excess & Five most favorable problems removed & $29.2$ [11.8, 47.7] \\ %
Gemma message-specific W$+$ excess & Five most favorable problems removed & $5.2$ [$-3.2$, 14.1] \\ %
Gemma message-specific W$+$ excess & Exclude Omni & $3.9$ [$-3.4$, 11.9] \\ %
Gemma fresh confirmation & Five most favorable problems removed & 21 versus bound 16.4 \\ %
Gemma fresh confirmation & Exclude SciBench; recompute BH & 19 versus bound 22.7 \\ %
Replay $-$ random removal & Five most favorable clusters removed & $2.35$ [1.91, 2.80] \\ %

\bottomrule
\end{tabular}
\par\smallskip{\footnotesize The Gemma selection margin and the global label excess survive these checks. Individual fresh confirmation is benchmark-sensitive, and the message-specific component is sensitive to concentrated problems. These are distinct claims and should not share one blanket robustness verdict.}
\end{table*}

\begin{table*}[t]
\centering\small
\caption{Project-wide multiplicity. One-sided $p$; adjusted values over the confirmatory family B$'$ (m = 11: the frozen pre-registered primaries plus the pre-specified headline tests). Holm controls the familywise error rate; Benjamini--Yekutieli (BY) controls the false discovery rate under arbitrary dependence.}
\label{tab:multiplicity}
\begin{tabular}{@{}l l r r r@{}}
\toprule
Hypothesis & Status & $p$ & Holm & BY \\
\midrule
Selector, new questions & pre-reg. & 0.808 & 1.0000 & 1.0000 \\
Learned ranker & pre-reg. & 0.459 & 1.0000 & 1.0000 \\
Zero-shot judge & pre-reg. & 0.585 & 1.0000 & 1.0000 \\
Fresh re-test, Gemma & pre-reg. & 0.0008 & 0.0055 & 0.0052 \\
Fresh re-test, OSS (amended) & pre-reg. & 0.566 & 1.0000 & 1.0000 \\
Latent-correct exclusion & pre-spec. & 0.0002 & 0.0018 & 0.0017 \\
Sufficiency test, Gemma & pre-spec. & 5.0e$-$5 & 0.0005 & 0.0017 \\
Sufficiency test, OSS & pre-spec. & 0.0165 & 0.0993 & 0.0916 \\
W$+$ excess, Gemma & pre-spec. & 0.0002 & 0.0018 & 0.0017 \\
W$+$ excess, OSS & pre-spec. & 0.259 & 1.0000 & 1.0000 \\
Replay-chosen $-$ random & pre-reg. ref. & 1.0e$-$4 & 0.0010 & 0.0017 \\
\bottomrule
\end{tabular}
\end{table*}

\clearpage
\section{J. A wrong-answer message that helps, and one that harms}
This illustration is selected post hoc from the public fresh-retest outputs, which exclude the human-audited problem set. The selection filters require a Gemma all-wrong pool with positive held-out $U$, outside Omni-MATH-2, with fresh-retest rows and one original message effect at least $+0.5$ and another at most $-0.5$. One pool qualifies. This is deliberately an informative example, not a representative sample.

\paragraph{Problem and messages.}
The MaScQA item G-META-17-54 \citep{zaki2023mascqa}, paraphrased, asks the maximum possible reduction when a 250\,mm slab is cold-rolled with 450\,mm-diameter rolls and a $10^\circ$ angle of bite. The reference accepts 6.60--7.40\,mm. The five proposals are 3.45, 17.45, 1.73, 11.03 and 39.67, all wrong under that reference.

Message A commits to 1.73 but states $\mu=\tan\alpha$ and $\Delta h_{\max}=R\tan^2\alpha$. Substituting $R=225$ and $\alpha=10^\circ$ yields 7.00, inside the reference range. It does not itself state that in-range value. Message B commits to 39.67. Its reasoning weighs a 3.42 computation against $\mu R=39.67$ and explicitly settles on 39.67. These summaries describe the generated texts; they do not independently validate every physical assumption or identify a unique useful formula in A.

\paragraph{Replay outcomes.}
In five original blocks, the integrator is correct in 5/10 full-pool calls, 0/5 calls without A and 5/5 calls without B. In 20 fresh blocks, the corresponding counts are 13/40, 0/20 and 19/20. Full-pool calls are shared across target comparisons.
The fresh effects are $+0.325$ [0.151, 0.499] for A and $-0.625$ [$-0.809$, $-0.441$] for B. These descriptive intervals are post hoc $t$ intervals over blocks, distinct from the problem-cluster intervals elsewhere. Individual directional exact tests give $p=0.002$ for A's helpful direction and $p=3.0\times10^{-6}$ for B's harmful direction. The original replay's correct outputs were 7.00; its wrong outputs were 3.42 or 39.67. This is consistent with different responses to the message contents, but removing entire messages does not show that a specific formula caused the benefit. Neither message is flagged as latent-correct by the reference-value classifier.

\section{K. Computational provenance and limits of reconstruction}
The controlled and fresh experiments have separate event sets. The new-question learned selector used a locked policy file, but no frozen hash receipt is retained; this differs from the hashed fresh-retest and format protocols. Major analyses were reimplemented by separate AI agents without access to the original analysis code. For the fresh re-test and format experiment, implementation followed frozen protocol text and synthetic tests before analyzing the new events. Post hoc reproductions re-derived estimates from raw data; some plans exposed prior aggregate results. These checks establish agreement between implementations, not independent experimental replication. Human judgments were supplied by two authors.

The original stable-label implementation was retained. The original permutation-null implementation is absent from the local archive, so it was reconstructed from its written specification; its ranges match the original printed ranges within one label, against a tolerance of three fixed before checking. The original C-comparator implementation is also absent. Of eight readings of its specification, two reproduce the archived comparator gains; they yield identical Gemma estimates and differ by 0.01 in two OSS interval endpoints. The comparison therefore has documented computational reconstruction uncertainty. It is not evidence that every historical source artifact is complete.

\paragraph{Code and data availability.}
The project website, code repository and Hugging Face dataset are linked on the first page. The public dataset at \url{https://huggingface.co/datasets/AgentsSci/AAAI_Wrong-but-Useful} contains the original measurement archive. The dataset revision checked for this release is \texttt{04a8aa45d4f7} (30 October 2026). That archive does not contain all additional experiments reported here. The reported analyses concern completed event sets; planned receiver-transfer and content-edit experiments are not used as supporting evidence.

\section{L. Core prompt templates}
The following core text is extracted from the configuration matching the recorded run metadata for both model families. Line wrapping is for display. Template variables denote supplied information; short benchmark-specific answer-format guidance is additional. These are prompts, not claims about the models' internal reasoning.
\subsection{Recruiter}
{\small\ttfamily\raggedright
You are coordinating a panel of independent reasoning agents who will each attempt the following problem from a DISTINCT, legitimate perspective:\par\smallskip
\$\{task\_description\}\par\smallskip
You will recruit \$\{cnt\_critic\_agents\} experts. Each expert must use a genuinely different reasoning lens (for example: structured decomposition; first-principles re-derivation; analogy / transfer to a known method; constraint / boundary / units analysis; estimation / sanity-bounding). Every expert must be a competent, good-faith solver — none is a strawman, and none should be told to be wrong. Diversity comes from the APPROACH, not from making mistakes.\par\smallskip
Recruit \$\{cnt\_critic\_agents\} experts whose perspectives are maximally complementary for THIS problem, so that even an imperfect hypothesis from one may contain a reasoning direction useful to a final integrator.\par\smallskip
\# Response Format Respond with a numbered list. Each line: <short role name> - <one-sentence role description naming the distinct reasoning lens>. 1. <name> - <description>. 2. <name> - <description>. ...\par\smallskip
}
\subsection{Hypothesizer}
{\small\ttfamily\raggedright
You are one of several INDEPENDENT reasoning agents working on the same problem. You have been assigned a role by the recruiter:\par\smallskip
Role: \$\{role\_description\}\par\smallskip
Problem: \$\{task\_description\}\par\smallskip
Make a serious, good-faith attempt to solve the problem FROM YOUR ASSIGNED PERSPECTIVE. You are working independently — you cannot see the other agents' work. Do not intentionally be wrong. Your proposed answer may be tentative, but it must be your genuine best attempt. Focus on contributing a distinct useful idea, decomposition, assumption, or check that a final integrator could build on.\par\smallskip
Respond in EXACTLY this format:\par\smallskip
Role: <your assigned role>\par\smallskip
Proposed final answer: \textbackslash{}boxed\{<your best concrete answer; required and must be concrete>\}\par\smallskip
Reasoning direction: <1-3 sentences describing the approach you took>\par\smallskip
Key idea: <the single most load-bearing insight in your attempt>\par\smallskip
Critical assumption: <the assumption your answer most depends on, which could be wrong>\par\smallskip
What another solver should check: <one concrete thing a verifier should check about your reasoning>\par\smallskip
Confidence / uncertainty: <low / medium / high, and one phrase on why>\par\smallskip
}
\subsection{Integrator}
{\small\ttfamily\raggedright
You are the final integrator. Several independent agents have each proposed a hypothesis for the following problem, from different perspectives:\par\smallskip
\$\{task\_description\}\par\smallskip
Below are their hypotheses. Some proposed answers may be wrong, but their reasoning directions, key ideas, or checks may still be useful. Do NOT blindly copy any proposed answer. Treat the hypotheses as evidence. Verify the reasoning yourself, reconcile disagreements, and synthesize your OWN final answer.\par\smallskip
\$\{hypotheses\_block\}\par\smallskip
Think step by step. You may adopt, combine, correct, or reject parts of any hypothesis, but justify your synthesis. Then give your final answer.\par\smallskip
Final answer: \textbackslash{}boxed\{<your synthesized final answer>\}\par\smallskip
}
\subsection{Evaluator}
{\small\ttfamily\raggedright
You are the Evaluator for a multi-agent reasoning system.\par\smallskip
Problem: \$\{task\_description\}\par\smallskip
MAS final submission: \$\{solution\}\par\smallskip
Ground truth / evaluator-only reference material: \$\{result\}\par\smallskip
Requirements:   - Decide whether the MAS final answer is mathematically equivalent to the ground truth.   - Focus on the final answer, not stylistic wording.   - The final answer may be numeric, symbolic, formulaic, set-valued, or conditional.   - Do not reveal, quote, restate, or hint at the ground truth in your response.   - Output a clear verdict label: `Verdict: PASS` or `Verdict: FAIL`.   - If the answer is equivalent, output `Correctness: 1`. Otherwise `Correctness: 0`.\par\smallskip
Output format: Verdict: PASS or FAIL Correctness: 0 or 1 Response: <brief non-leaking note>\par\smallskip
}
\subsection{Independent solver}
{\small\ttfamily\raggedright
Solve the following problem independently and carefully.\par\smallskip
\$\{task\_description\}\par\smallskip
\$\{domain\_guidance\}\par\smallskip
Think step by step, then give your final answer.\par\smallskip
Final answer: \textbackslash boxed\{<your final answer>\}\par
}

\end{document}